\documentclass[letterpaper, 10 pt, conference]{ieeeconf}  % Comment this line out if you need a4paper

\IEEEoverridecommandlockouts
\usepackage{cite}
\usepackage{amsmath,amssymb,amsfonts,mathtools}
\usepackage{algorithm, algorithmic}
\usepackage{graphicx}
\usepackage{textcomp}
\usepackage{xcolor}
\usepackage{bm}
\usepackage{pstool}
\usepackage{relsize}
\usepackage{paralist}
\usepackage{siunitx}
\usepackage{booktabs}
\usepackage{colortbl}
\usepackage{cases}
\usepackage{blindtext}
\usepackage{etoolbox} % to hook into `figure` and reset the mysubfig counter
\usepackage{multirow}
\usepackage{scalerel}
\usepackage{letltxmacro}
\renewcommand*{\mathellipsis}{%
  \mathinner{{\ldotp}{\ldotp}{\ldotp}}%
}
\makeatletter
\@ifdefinable{\org@ldots}{%
  \LetLtxMacro\org@ldots\ldots
  \DeclareRobustCommand*{\ldots}{%
    \ifmmode
      \expandafter\my@ldots
    \else
      \expandafter\textellipsis
    \fi
  }%
}
\newcommand*{\neghalfmskip}{%
  \nonscript\mskip-.5\muexpr\thinmuskip\relax%
}
\newcommand*{\my@ldots}{%
  \mathellipsis
  \@ifnextchar,\neghalfmskip{%
  \@ifnextchar:\neghalfmskip{%
  \@ifnextchar;\neghalfmskip{%
  \@ifnextchar.\neghalfmskip{%
  \@ifnextchar!\neghalfmskip{%
  \@ifnextchar?\neghalfmskip{%
  \@ifnextchar){\mskip-.5\muexpr\thinmuskip\relax}{% negative kerning
  }}}}}}}%
}
\makeatother

\def\matr#1{\bm{\uppercase{#1}}}
\def\vect#1{\bm{\lowercase{#1}}}

\def\set#1{\mathcal{\uppercase{#1}}}
\def\tuple#1{\mathtt{#1}}
\def\lst#1{\mathsf{#1}}
\def\testc{\tuple{C}}
\def\testcs{\tuple{S}}
\def\locat{{M}}
\def\stateX{\vect{x}}
\def\outputY{\vect{y}}
\def\inputU{\vect{u}}
\def\setStateX{\set{X}}

\def\reachX{\set{X}}
\def\reachY{\set{Y}}
\def\iParent#1{_{#1}}
\def\iChildCom#1{_{#1}}
\def\iChildCase#1{}
\def\iChildSec#1{^{\scalebox{0.5}{$($}#1\scalebox{0.5}{$)$}}}
\def\iInitial{_0}
\def\iInitialS{[0]}

\def\iEnd{_{\star}}

\def\funcName#1{\text{\tt{#1}}}
\def\iNorm{_{*}}

\def\iMatrix#1{[{#1}]}

\newcommand{\mycomment}[1]{}

\newlength{\bracewidth}

\newtheorem{defi}{Definition}

\def\BibTeX{{\rm B\kern-.05em{\sc i\kern-.025em b}\kern-.08em
    T\kern-.1667em\lower.7ex\hbox{E}\kern-.125emX}}

\begin{document}

\title{\LARGE \textbf{Efficiently Obtaining Reachset Conformance for the Formal Analysis \\ of Robotic Contact Tasks}}

\author{Chencheng Tang and Matthias Althoff% <-this % stops a space
\thanks{
Authors are with the School of Computation, Information and Technology, Technical University of Munich, 85748 Garching, Germany.
Email: {\tt\small \{chencheng.tang,althoff\}@tum.de}}%
}

\maketitle
%%%%%%%%%%%%%%%%%%%%%%%%%%%%%%%%%%%%%%%%%%%%%%%%%%%%%%%%%%%%%%%%%%%%%%%%%%%%%%%%
\begin{abstract}
Formal verification of robotic tasks requires a simple yet conformant model of the used robot.
We present the first work on generating reachset conformant models for robotic contact tasks considering hybrid (mixed continuous and discrete) dynamics.
Reachset conformance requires that the set of reachable outputs of the abstract model encloses all previous measurements
to transfer safety properties.  
Aiming for industrial applications, we describe the system using a simple hybrid automaton with linear dynamics.
We inject non-determinism into the continuous dynamics and the discrete transitions,
and we optimally identify
all model parameters together with the non-determinism required to capture the recorded behaviors.
Using two 3-DOF robots, we show that our approach can effectively generate models to capture uncertainties in system behavior
and substantially reduce the required testing effort in industrial applications.
\end{abstract}

%%%%%%%%%%%%%%%%%%%%%%%%%%%%%%%%%%%%%%%%%%%%%%%%%%%%%%%%%%%%%%%%%%%%%%%%%%%%%%%%
\section{Introduction}
Contact tasks \cite{ControlRobotsContact2009}
represent an increasingly large share of robotic applications \cite{suomalainenSurveyRobotManipulation2022,IndustrialRoboticsInsightsa}. %, such as material-removing and assembling. 
Verifying specifications of a contact task using traditional testing methods can take several hours even after minor modifications to the system \mbox{(e.g., \cite[Sec. \uppercase\expandafter{\romannumeral5\relax}]{kirschnerISOTS150662022})}. 
To efficiently guarantee the safety and success of contact tasks, reachability analysis has been employed \cite{tangFormalVerificationRobotic2023},
which computes the set of states or outputs that are reachable by a system model and captures all possible behaviors for formal verification.
However, we still require appropriate models to capture all possible behaviors of a real robotic system
to transfer verification results.

To address the aforementioned problem,
we synthesize models of contact tasks that are reachset conformant.
\textit{Reachset conformance} requires that all measurements of an implementation are enclosed by the set of reachable outputs of the abstract model \cite{roehmReachsetConformanceTesting2016}.
Reachset conformance is necessary and sufficient for transferring safety properties \cite{roehmReachsetConformanceAutomatic2022};
alternatives, such as \textit{trace conformance} \cite{dangModelBasedTestingHybrid2012} and \textit{simulation relations} \cite{haghverdiBisimulationRelationsDynamical2005}, are not considered,
because these properties are harder to achieve for robotic systems considering arbitrary disturbances and sensor noise \cite[p. 2]{liuGuaranteesRealRobotic2023}.

To over-approximately capture the errors between the behavior of a model and its complex real counterpart,
non-determinism is injected.
For example, \cite{lutzowReachsetConformantIdentificationNonlinear2024,liuReachsetConformanceForward2018} consider uncertainties in the initial state and the input,
while \cite{liuGuaranteesRealRobotic2023,kochdumperEstablishingReachsetConformance2020} add errors to the flow and output function.
In the field of robotics,
reachset conformance has provided promising results in dynamics modeling\cite{liuReachsetConformanceForward2018}, human-robot interaction\cite{liuOnlineVerificationImpactForceLimiting2021}, and position control \cite{liuGuaranteesRealRobotic2023},
but hybrid dynamics have not yet been considered.

Tools for formal analysis \cite{althoffIntroductionCORA20152015,chenFlowAnalyzerNonlinear2013,bakHyLAAToolComputing2017,frehseSpaceExScalableVerification2011,gurungParallelReachabilityAnalysis2016}
commonly require simple models,
and linear dynamics are usually preferred for better computation efficiency in real-world robotic applications \cite{liuReachsetConformanceForward2018, liuOnlineVerificationImpactForceLimiting2021,liuGuaranteesRealRobotic2023}.
Consequently,
the trade-off between the simplicity of the model and the required non-determinism is crucial;
an overly conservative model obviously cannot provide accurate verification results.
\textit{Conformance synthesis} addresses this by
automatically determining the non-determinism required for establishing reachset conformance.
Recent works have shown that the synthesis problem can be optimally solved with linear programming when sets are represented by zonotopes \cite{liuGuaranteesRealRobotic2023, lutzowReachsetConformantIdentificationNonlinear2024}.
For hybrid systems, these techniques have not been examined yet,
and conformance synthesis has only been done in a heuristic way for analog circuits \cite{kochdumperEstablishingReachsetConformance2020}, 
which uses a precise simulation of the real system to determine the process error and then encloses real measurements by bloating the reachable sets
-- a sufficiently precise model is usually not available for contact tasks with complex dynamics.

We present the first work in optimally synthesizing reachset conformant models for hybrid systems.
Also, it is the first time that reachset conformance is established for robotic systems with hybrid dynamics considered.
We show and examine our approach using a constrained collision scenario,
which is essential in contact tasks and is therefore often used in industrial standards (e.g., \cite{ISOTS15066})
and applied research (e.g.,\cite{haddadinRoleRobotMass2008,achhammerImprovementModelmediatedTeleoperation2010,chenRoboticGrindingBlisk2019}).
Non-determinism is introduced to capture not only the uncertainties in the continuous process
but also the errors introduced by discrete transitions.
All model parameters, including the transition conditions,
are identified by optimizing the non-determinism.

We first introduce the preliminaries and formalize the problem in Sec. \ref{sec_pre}.
Contact task scenarios are modeled in Sec. \ref{sec_sys}.
We introduce the approach for synthesizing reachset conformant models in Sec. \ref{sec_syn}.
Our approach is evaluated on two 3-DOF planar robots in various testing conditions in \mbox{Sec. \ref{sec_exp}}.

%%%%%%%%%%%%%%%%%%%%%%%%%%%%%%%%%%%%%%%%%%%%%%%%%%%%%%%%%%%%%%%%%%%%%%%%%%%%%%%%
\section{Preliminaries and Problem Statement} \label{sec_pre}
This section poses the problem after recalling some preliminaries.
We denote vectors by bold lowercase letters (e.g., $\vect{a}$), matrices by bold uppercase letters (e.g., $\matr{a}$), lists by sans-serif font (e.g., $\lst{A} = (a,b)$), tuples by typewriter font (e.g., $\tuple{A} = \langle a,\matr{B} \rangle$), and sets by calligraphic font (e.g., $\set{a}$).
We use SI units unless stated otherwise.

\subsection{Preliminaries}\label{subsec_pre}
In this work, we represent sets by zonotopes \cite{kuhnRigorouslyComputedOrbits1998}:
\begin{defi}[Zonotope]\label{def_zonotope}
  Given a center $\vect{c} \in \mathbb{R}^n$ and a generator matrix $\matr{G}=[\vect{g}_{1},\ldots,\vect{g}_{\eta}] \in \mathbb{R}^{n \times \eta}$,
  a zonotope is
  \begin{equation*}
        \set{Z} =
        \langle\vect{c},\matr{G}\rangle :=       \bigg\{ \stateX = \vect{c} + \sum_{i = 1}^{\eta} \beta_i\,\vect{g}_{i} \,\bigg|\, \beta_i\in[-1,1] \bigg\}.
  \end{equation*}
\end{defi}
On zonotopes, many operations can be exactly and efficiently computed \cite{althoffSetPropagationTechniques2021},
such as Minkowski addition and linear transformation ($\set{Z}_a = \langle\vect{c}_a, \matr{G}_a\rangle$, $\set{Z}_b = \langle\vect{c}_b, \matr{G}_b\rangle$):
\begin{align}
  &\set{Z}_a \mkern-2mu \oplus \mkern-2mu \set{Z}_b  \mkern-2mu :=  \mkern-2mu \left\{\vect{a}+\vect{b} \,\middle|\, \vect{a}\in\set{Z}_a,\vect{b}\in\set{Z}_b\right\} \mkern-2mu  = \mkern-2mu  \langle\vect{c}_a+\vect{c}_b,[\matr{G}_a \,\matr{G}_b]\rangle, \nonumber \\
  &\matr{M}\set{Z} := \left\{\matr{M}\vect{x} \,\middle|\, \vect{x}\in\set{Z}\right\} = \langle \matr{M}\vect{c}, \matr{M}\matr{G} \rangle.  \nonumber
\end{align}
We evaluate the size of a zonotope using its interval norm $\|\set{Z}\|_I = \sum_{i=1}^{\eta}{\| \vect{g}_i \|_1}$ (see \cite[Def. 5]{lutzowReachsetConformantIdentificationNonlinear2024}).

To describe the hybrid dynamics, we formalize contact tasks as hybrid automata similar to \cite{kochdumperReachabilityAnalysisHybrid2020}:
\begin{defi}[Hybrid Automaton]\label{def_ha}
  A hybrid automaton $\tuple{H}$ has multiple discrete states referred to as locations or modes,
  where $\locat_{m}$ represents the $m$-th location.
  With non-determinism injected, $\tuple{H}$ consists of:
  \begin{itemize}
    \item Flow of each location $\dot{\stateX} \in f_m(\stateX,\inputU) \oplus \set{W}_m$ describing the continuous dynamics with flow function $f_m$, continuous state $\stateX \in \mathbb{R}^n$, \mbox{input $\inputU \in \mathbb{R}^\zeta$}, and process disturbance set $\set{W}_m$.
    \item Output of each location ${\outputY} \in l_m(\stateX,\inputU) \oplus \set{V}_m, \outputY \in \mathbb{R}^o$ with output function $l_m$ and measurement error set $\set{V}_m$.
    \item Invariant sets of each location $\set{I}_m\subset\mathbb{R}^n$ describing the region where the flow function $f_m$ is valid.
    \item A list of discrete transitions, where the $q$-th transition $\tuple{Q}\iParent{q}=\langle\set{G}\iChildCom{q}, r\iChildCom{q}(\stateX), \set{Q}_q, s\iChildCom{q},d\iChildCom{q}\rangle$
    contains a guard set \mbox{$\set{G}\iChildCom{q}\subset\mathbb{R}^n$},
    a reset map $\stateX' \in r\iChildCom{q}(\stateX) \oplus \set{Q}_q$ that defines the continuous state after transition $\stateX'$ with reset function $r\iChildCom{q}(\stateX)$ as well as transition disturbance set $\set{Q}_q$,
    and $s\iChildCom{q}, d\iChildCom{q}$, which are indices of the source and destination location, respectively.
  \end{itemize}
\end{defi}

\begin{figure}[t]
  \vspace{0em}
  \centering
  \includegraphics[width=0.9\columnwidth]{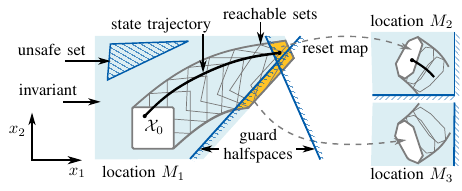}
  \caption{State evolution in a hybrid automaton.
  }
  \vspace{0em}
  \label{fig_hasemantics}
\end{figure}

The evolution of a hybrid automaton $\tuple{H}$ is illustrated in Fig.~\ref{fig_hasemantics} and is informally described as follows:
Given an initial location with index ${m\iInitial}$ and an input sequence $\inputU[\cdot]$,
where we use $[\cdot]$ to denote a sequence and $[k]$ to denote the $k$-th time step,
the continuous state starts from initial state $\stateX\iInitial \in \setStateX\iInitial$ and evolves following the flow function $f_{m\iInitial}(\stateX,\inputU)$
with a disturbance \mbox{$\vect{w}(\cdot) \in \set{W}_{m\iInitial}$}.
The output is the result of the output function $l_{m\iInitial}(\stateX,\inputU)$ with an error $\vect{v}(\cdot) \in \set{V}_{m\iInitial}$.
When $\stateX$ is within the guard set of a transition,
the state may transit to the corresponding target location in zero time,
but it may also stay in the same location until leaving the invariant set $\set{I}_{m_0}$.
In case several guard sets are hit at the same time, the transition is chosen non-deterministically.
After a transition $\tuple{Q}\iParent{q}$, the continuous state is updated by the reset function $r\iChildCom{q}(\stateX)$ with a disturbance $\vect{q} \in \set{Q}_{q}$ added,
and the result becomes the initial state in the new location where the evolution continues.
Accordingly, we denote a possible output trajectory of the system
by $\xi(t,\stateX\iInitial,m\iInitial,\inputU[\cdot])$.

%%%%%%%%%%%%%%%%%%%%%%%%%%%%%%%%%%%%%%%%%%%%%%%%%%%%%%%%%%%%%%%%%%%%%%%%%%%%%%%%
\subsection{Problem Statement} \label{subsec_pro}
We model a contact task as a hybrid automaton with linear dynamics,
which is defined by a vector of model parameters $\vect{p}$
and lists $\lst{W}, \lst{V}, \lst{Q}$ respectively storing $\set{W}_m, \set{V}_m$ of each location and $\set{Q}_q$ of each transition.
For reachset conformance, we check whether the reachable sets of the output of $\tuple{H}$ enclose the measurements collected from the real implementation.
\begin{defi}[Reachable Set]
  Given a hybrid automaton $\tuple{H}$ defined in Def. \ref{def_ha},
  an initial location indexed as $m\iInitial$,
  an initial set $\setStateX\iInitial$,
  and an input sequence $\inputU[\cdot]$ assuming zero-order hold,
  the reachable set of the output at time $t$ is:
  \begin{equation*}
    \reachY(t,\setStateX\iInitial, m\iInitial,\inputU[\cdot]) =
    \bigl\{\xi(t,\stateX\iInitial,m\iInitial,\inputU[\cdot]) \,\big|\,
    \stateX\iInitial \in \setStateX\iInitial \bigr\},
  \end{equation*}
  and the reachable set of the continuous state $\reachX$ is a special case with $\outputY=\stateX$.
  %We denote the part of $\reachX$ in location $\locat_m$ by $\reachX_m$ and the associated output set by $\reachY_m$.
\end{defi}
\begin{defi}[Test Case] \label{def_tc}
  A test case used in conformance checking $\testc = \langle \stateX\iInitial\iChildCase{}, m\iInitial\iChildCase{}, \inputU\iChildCase{}[\cdot], \outputY\iChildCase{}[\cdot], t\iChildCase{}[\cdot]\rangle$
  consists of an initial state $\stateX\iInitial\iChildCase{}$,
  an initial location indexed as $m\iInitial\iChildCase{}$,
  and $k\iEnd$ data points consisting of the inputs $\inputU[\cdot]$ and measured outputs $\outputY\iChildCase{}[\cdot]$ collected at timestamps $t\iChildCase{}[\cdot]$.
\end{defi}

We consider one test case $\testc$ for cleaner notation,
as our approach works analogously for any number of test cases.
The problem is to find the optimal values of $\vect{p},\lst{W},\lst{V},\lst{Q}$
resulting in the smallest reachable sets while ensuring reachset conformance:
\vspace{0em}
\begin{subequations} \label{eq_pro}
  \begin{align}
    &\min_{\vect{p},\lst{W},\lst{V},\lst{Q}} \funcName{cost}(\testc, \vect{p},\lst{W},\lst{V},\lst{Q}) \label{eq_pro_1} \\
    &\text{s.t.} \mkern10mu \forall k:\outputY\iChildCase{}[k] \in {\reachY(t\iChildCase{}[k],\stateX\iInitial\iChildCase{},m\iInitial\iChildCase{},\inputU\iChildCase{}[\cdot]).}\label{eq_pro_3}%_{\reachY\iChildCase{}[k]}. 
  \end{align} 
\end{subequations}
The size of sets is evaluated using a cost function \funcName{cost}, which is presented in Sec. \ref{subsec_cost}.

%%%%%%%%%%%%%%%%%%%%%%%%%%%%%%%%%%%%%%%%%%%%%%%%%%%%%%%%%%%%%%%%%%%%%%%%%%%%%%%%

\section{System Modeling} \label{sec_sys}
\begin{figure}[t]
  \centering
  \includegraphics[width=1\columnwidth]{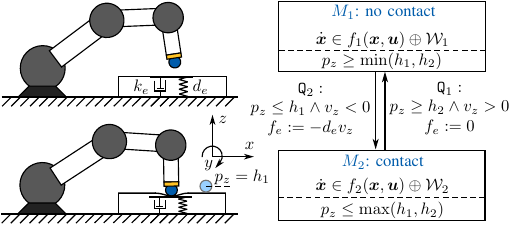}
  \caption{The hybrid automaton of the representative scenario.
    The end effector (blue) is shaped like a sphere whose center locates the tool center point.
    A force sensor (yellow) measures the external force.
    A location is represented by a box with the flow function above the dotted line and the invariant below.
    Guards and reset functions are presented next to the transition arrows.
  }
  \label{fig_scene}
  \vspace{0em}
\end{figure}

We consider a representative contact scenario,
where a robot hits a surface during task execution,
as shown by the hybrid automaton in \mbox{Fig. \ref{fig_scene}} with two locations:
\textit{no contact} $\locat_1$ and \textit{contact} $\locat_2$.
Transitions happen when the robot hits the surface at height $h_1$ and leaves the surface at $h_2$.
To describe the system dynamics, we define the state vector
\mbox{$\stateX = 
\begin{bmatrix}
  p_z &
  v_z &
  f_e &
  p_x &
  \theta_y
\end{bmatrix}^\mathsf{T}$},
where $p_x, \theta_y, p_z$ respectively represent the positions in spatial dimensions $x, y, z$,
and we denote the corresponding velocities by $v_x, \omega_y, v_z$.
The output vector is \mbox{$\outputY = \begin{bmatrix}
  p_z &
  f_e &
  p_x &
  \theta_y
\end{bmatrix}^\mathsf{T},
$}
and the input vector
\mbox{
$\inputU = 
\begin{bmatrix}
  {p}_{z,d} &
  {v}_{z,d} &
  {v}_{x,d} &
  {\omega}_{y,d}
\end{bmatrix}^\mathsf{T}$}
consists of the commands from the input trajectory, which are denoted by subscript $d$.

The nominal flow function $f(\stateX,\inputU)$ is formulated as follows:
In the vertical direction $z$,
the dynamics is the result of an actuation force ${f}_r$ and an external force ${f}_e$ applied to an effective mass $m_r$ \cite{khatibInertialPropertiesRobotic1995}:
\begin{equation} \label{eq_dynZ}
  \dot{p}_z = v_z,\, \dot{v}_z = \frac{{f}_r+{f}_e}{m_r},
\end{equation}
where the robot is under Cartesian admittance control \cite{albu-schafferCartesianImpedanceControl2002} to behave compliantly like a mass-spring-damper system with spring factor $k_r$ and damping factor $d_r$:
\begin{equation} \label{eq_fr}
  f_r = k_r({s}_{z,d}-{{s}}_z)+d_r({v}_{z,d}-{v}_z).
\end{equation}
The normal force ${f}_e$ is modeled using the Kelvin-Voigt model,
which is widely used for its simplicity and effectivness \cite{gilardiLiteratureSurveyContact2002}:
\begin{subequations} \label{eq_fe}
  \begin{numcases}{f_e =}
    0 & if $m=1$ \label{eq_fe_nc} \\
    -k_e (p_z-h_1) - d_e v_z & otherwise, \label{eq_fe_wc}
  \end{numcases}
\end{subequations}
where $k_e, d_e$ are respectively the stiffness and damping coefficients.
For other spatial dimensions $x$ and $y$, position control is applied with external forces ignored:
\begin{equation}
  \dot{p}_x = v_x = {v}_{x,d},\, \dot{\theta}_y = \omega_y = {\omega}_{y,d}.
\end{equation}
Next, we identify parameter values and add non-determinism to this model to obtain a reachset conformant model.

%%%%%%%%%%%%%%%%%%%%%%%%%%%%%%%%%%%%%%%%%%%%%%%%%%%%%%%%%%%%%%%%%%%%%%%%%%%%%%%%
\section{Synthesizing Reachset Conformant Models} \label{sec_syn}

\begin{figure*}[t]
  \centering
  \includegraphics[width=1.95\columnwidth]{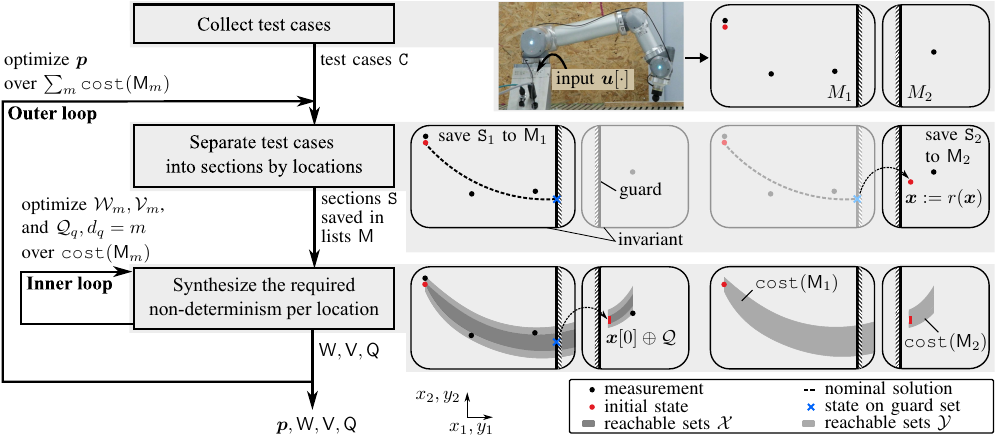}
  \caption{Concept for synthesizing reachset conformant models for robotic contact tasks.
  }
  \label{fig_ident}
  \vspace{0em}
\end{figure*}

We solve problem \eqref{eq_pro} and optimally establish reachset conformance as illustrated in Fig. \ref{fig_ident}:
Based on an initial guess of the parameters $\vect{p}$,
the test cases are separated by the locations.
The data of each location is used in an inner loop
to optimize $\lst{W},\lst{V},\lst{Q}$ using linear programming.
The outer loop evaluates the resulting cost and optimizes $\vect{p}$ using nonlinear programming.
Below, we explain the procedure in detail.

\subsection{Test Case Processing for Hybrid Systems} \label{subsec_testCases}
As mentioned above, we separate test cases by locations into sections.

\begin{defi}[Test Case Section] \label{def_tcs}
  A test case $\testc$ is split by the locations;
  each section $\testcs = \langle \stateX[0], q, \inputU\iChildCase{}[\cdot], \outputY\iChildCase{}[\cdot], t\iChildCase{}[\cdot], \stateX[j\iEnd]\rangle$ starts from the initial state $\stateX\iInitialS$ at time $t\iInitialS$,
  when the source transition (index $q$) is triggered,
  and contains the sequences \mbox{$\inputU[\cdot], \outputY[\cdot], t[\cdot]$} from $\testc$ for the times
  until $t[j\iEnd]$, when the nominal state $\stateX[j\iEnd]$ triggers the transition to the next section.
\end{defi}

\begin{figure}[t]
  \centering
  \includegraphics[width=1\columnwidth]{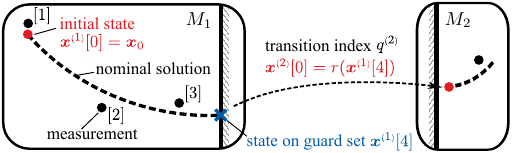}
  \caption{Separating a test case $\testc$ by locations into sections. We assume $\outputY = \stateX$ to simplify the visualization,
  and we denote the elements of the $s$-th section $\testcs_{s}$ using the superscript $(s)$.
  }
  \label{fig_data}
  \vspace{0em}
\end{figure}

The preprocessing is repeated for each test case as shown in Fig. \ref{fig_data}.
Given a test case $\tuple{C}$ starting in location $\locat_1$,
we compute the nominal state trajectory $\stateX\iNorm$ using \eqref{eq_compute2} starting from the initial state $\stateX\iInitial$ until it intersects a guard set.
The data before the transition time $t\iChildSec{1}[4]$ composes the first data section $\testcs_1$,
including the initial values (red in Fig. \ref{fig_data}), the data from $\tuple{C}$ (black in \mbox{Fig. \ref{fig_data}}),
and the state associated with the transition (blue in Fig. \ref{fig_data}).
We save $\testcs_1$ to a list $\lst{M}_1$, which stores all data belonging to location $\locat_1$ for conformance synthesis. 
Then,
we employ the reset function to obtain the nominal initial state $\stateX\iChildSec{2}\iInitial$ of the next section $\testcs_2$,
and we save the index of the triggered transition $q\iChildSec{2}$ to $\testcs_2$.
By repeating the above procedure until the end of each test case,
we obtain a list $\lst{M}_m$ which stores all associated sections for each location $\locat_m$.

%%%%%%%%%%%%%%%%%%%%%%%%%%%%%%%%%%%%%%%%%%%%%%%%%%%%%%%%%%%%%%%%%%%%%%%%%%%%%%%%
\subsection{Decoupling Discrete States in Conformance Synthesis} \label{subsec_decoupl}

The conformance synthesis problem \eqref{eq_pro} is computationally challenging,
as the reachable sets in \eqref{eq_pro_3} depend on every traversed location.
We propose to identify each location independently.
To simplify the notation,
we do not mention the section and omit the superscripts when describing the synthesis of the reachset conformant model for a specific section $\testcs_s$ in $\locat_m$.

Given the initial state $\stateX_0$ of a test case,
the initial set of a section is the set propagation at time $t[0]$,
i.e., $\reachX(t\iInitialS, \stateX\iInitial, m\iInitial,\inputU[\cdot])$.
We under-approximate $\reachX(\ldots)$ using its subset $\stateX\iInitialS \oplus \set{Q}_{q}$ to simplify the computation in Sec. \ref{subsec_reachsetLinear},
and we rewrite the reachset conformance problem \eqref{eq_pro_3} for each section as
\begin{equation} \label{eq_yinY}
  \forall j: \outputY[j] \in \underbrace{{\reachY}\big(t[j], \stateX\iInitialS \oplus \set{Q}_{q}, m,\inputU[\cdot]\big)}_{\subseteq{\reachY}\big(t[j], \stateX_0, m_0,\inputU[\cdot]\big)}.
\end{equation}
The reachable set in \eqref{eq_yinY} is computed
as presented next in Sec. \ref{subsec_reachsetLinear}
without considering the invariant set.
This simplification is sound,
because if the system is reachset conformant even beyond the invariant set,
it is certainly reachset conformant within the invariant set.
Consequently,
we formulate the containment problem \eqref{eq_yinY} of each section \mbox{$\testcs_s\in\lst{M}_m$} as linear constraints to synthesize location $\locat_m$.

\subsection{Reachset Conformance as Linear Constraints} \label{subsec_reachsetLinear}
For each test case section, we compute ${\reachY}$ as follows with
flow function $f(\stateX,\inputU) = \matr{A}\stateX + \matr{B}\inputU$
and output function $l(\stateX,\inputU) = \matr{C}\stateX + \matr{D}\inputU$ \cite[Prop. 1]{lutzowReachsetConformantIdentificationNonlinear2024}:
\begin{equation} \label{eq_compute2}
  \begin{split}
  %&(\prod_{i=0}^{j} \bar{\matr{A}}\iMatrix{j-i})\set{Q} \oplus \bar{\set{W}}[j] \oplus \bigoplus_{k=0}^{j-1} (\prod_{i=0}^{j-k-1} \bar{\matr{A}}\iMatrix{j-i})\bar{\set{W}}[k]\\
  & {\reachY}[j] = \matr{C}{\reachX}[j] \oplus \matr{D}\inputU[j] \oplus \set{V}, \\
  & {\reachX}[j] = \stateX\iNorm[j] \oplus \bar{\matr{A}}(0,j)\set{Q}_q \oplus \bigoplus_{i=0}^{j-1} \bar{\matr{A}}(i+1,j)\bar{\set{W}}[i], \\
  & \stateX\iNorm[j] = \bar{\matr{A}}(0,j) \stateX\iInitialS + \sum_{i=0}^{j-1} \bar{\matr{A}}(i+1,j)\bar{\matr{B}}\iMatrix{i}\inputU[i],
  \end{split}
\end{equation}
where with $\Delta t[j] = t[j+1] - t[j]$,
\begin{align}
  & \bar{\matr{A}}(i,j) = e^{\matr{A} (t[j] - t[i])}, \bar{\matr{B}}\iMatrix{j} = \int_{0}^{\Delta t[j]}\hspace{-0.5em}e^{\matr{A}({\Delta t[j]}-\tau)}d \tau \, \matr{B}, \nonumber \\
  & \bar{\set{W}}[j] = \biggl\{\int_{0}^{\Delta t[j]}\hspace{-0.5em}e^{\matr{A}({\Delta t[j]}-\tau)} \vect{w} d \tau \,\bigg|\, \vect{w} \in \set{W}\biggr\}. \label{eq_compute3}
\end{align}

To factor out $\set{W}$ from \eqref{eq_compute3},
we consider a constant $\vect{w}$ during each time step $[t[j], t[j+1]]$ as in \cite[Prop. 1]{liuReachsetConformanceForward2018}
and under-approximate \eqref{eq_compute2} with
\begin{equation}
  \bar{\set{W}}[j] \supseteq \int_{0}^{\Delta t[j]}\hspace{-0.5em}e^{\matr{A}({\Delta t[j]}-\tau)}d \tau \, \set{W} = {\matr{E}_w[j]} \, \set{W}. \label{eq_compute4}
\end{equation}
Accordingly, we rewrite the terms of $\set{Q}$ and $\set{W}$ in \eqref{eq_compute2} as:
\begin{subequations} \label{eq_compute5}
\begin{align}
  &{\matr{E}_1[j]}\set{Q}_q = \bar{\matr{A}}(0,j)\set{Q}_q, \\
  & \matr{E}_2[j]\set{W} = \bigoplus_{i=0}^{j-1} \bar{\matr{A}}(i+1,j)\matr{E}_w[i]\set{W}. \label{eq_compute5_b}
\end{align}
\end{subequations}

We omit the dependency of \eqref{eq_yinY} on the initial state and the input trajectory
by subtracting the nominal solution \mbox{$\outputY\iNorm[j] = \matr{C}(\stateX\iNorm[j]) + \matr{D}(\inputU[j])$} from both sides \cite[Sec. \uppercase\expandafter{\romannumeral3\relax}]{liuGuaranteesRealRobotic2023}:
\begin{equation} \label{eq_linearContain_1}
  \forall j: \outputY[j] - \outputY\iNorm[j]
  \in \underbrace{\matr{C}\matr{E}_1[j]\set{Q}_q \oplus \matr{C}\matr{E}_2[j]\set{W} \oplus \set{V}}_{\reachY[j]-\outputY\iNorm[j] \; \text{with} \; \eqref{eq_compute2}, \eqref{eq_compute5}}.
\end{equation}
To fulfill the assumption $\stateX\iInitialS \oplus \set{Q}_{q} \subseteq \reachX(t\iInitialS, \stateX\iInitial, m\iInitial,\inputU[\cdot])$,
we add the following constraint:
\begin{equation} \label{eq_linearContain_2}
  \underbrace{\stateX[j\iEnd] - \stateX\iNorm[j\iEnd]}_{=\vect{0}\;\text{using Sec. \ref{subsec_testCases}}} \in \underbrace{\matr{E}_1[j\iEnd]\set{Q}_q \oplus \matr{E}_2[j\iEnd]\set{W}}_{{\reachX}[j\iEnd] - \stateX\iNorm[j\iEnd]  \; \text{with} \; \eqref{eq_compute2}, \eqref{eq_compute5}}.
\end{equation}
The constraints in \eqref{eq_linearContain_1} and \eqref{eq_linearContain_2} for reachset conformance are formulated as linear inequalities
using the halfspace representation \mbox{\cite[Thm. 1]{liuGuaranteesRealRobotic2023}}
or the generator representation \mbox{\cite[Thm. 3]{lutzowReachsetConformantIdentificationNonlinear2024}} of zonotopes.
We restrict the sets $\set{W},\set{V},\set{Q}_q$ to be zonotopes of the form:
\begin{align}
  & \set{W} = \langle \vect{c}_W, \matr{G}_W \, \funcName{diag}(\vect{\alpha}_W) \rangle,
  \vect{\alpha}_W \in \mathbb{R}^{\eta_W}_{\geq 0},  \nonumber \\
  & \set{V} = \langle \vect{c}_V, \matr{G}_V \, \funcName{diag}(\vect{\alpha}_V) \rangle, 
  \vect{\alpha}_V \in \mathbb{R}^{\eta_V}_{\geq 0},  \nonumber \\
  & \set{Q}_q = \langle \vect{c}_{Q_q}, \matr{G}_{Q_q} \, \funcName{diag}(\vect{\alpha}_{Q_q}) \rangle,
  \vect{\alpha}_{Q_q} \in \mathbb{R}^{\eta_{Q_q}}_{\geq 0}. \nonumber
\end{align}
The generator templates \mbox{$\matr{G}_W \in \mathbb{R}^{n \times \eta_W}$}, \mbox{$\matr{G}_V \in \mathbb{R}^{o \times \eta_V}$}, $\matr{G}_{Q_q} \in \mathbb{R}^{n \times \eta_{Q_q}}$
define the structure of the sets,
and the numbers of generators $\eta_W, \eta_V, \eta_{Q_q}$ determine the computation complexity;
often, using identity matrices already leads to good results.
Accordingly, we optimize $\vect{c}_W, \vect{\alpha}_W, \vect{c}_V, \vect{\alpha}_V$,
and all $\vect{c}_{Q_q}, \vect{\alpha}_{Q_q}$ with \mbox{$d_q = m$}
to synthesize a location with linear programming.

The under-approximation \eqref{eq_compute4} is close to its exact counterpart when the time steps are small.
However, 
following the complexity computation in \mbox{\cite[Prop. 2]{lutzowReachsetConformantIdentificationNonlinear2024}},
the halfspace enclosure method \mbox{\cite[Thm. 1]{liuGuaranteesRealRobotic2023}} falls short when the number of time steps $j\iEnd$ is large,
as the growing number of generators in \eqref{eq_compute5_b} leads to $\mathcal{O}({j\iEnd}^{o})$ constraints from \eqref{eq_linearContain_1}. % due to the exponentially increasing number of linear constraints.
A further under-approximation is useful,
where we consider a constant $\vect{w}$ during $[t[0], t[j]]$:
\begin{equation} \label{eq_underapproxW}
  \matr{E}_2[j]\set{W} = \left(\sum_{i=0}^{j-1} \bar{\matr{A}}(i+1,j)\matr{E}_w[i]\right)\set{W},
\end{equation}
which has a fixed number of $\eta_W$ generators thus decreases the number of constraints to $\mathcal{O}({j\iEnd} \, \gamma^{o}), \gamma \leq \frac{(\eta_W+\eta_V+\eta_{Q_q})e}{o-1}$.

\subsection{Evaluating the Costs of Reachable Sets} \label{subsec_cost}

For the cost function \eqref{eq_pro_1},
we use the interval norm of zonotopes
as in \cite[Lem. 1]{lutzowReachsetConformantIdentificationNonlinear2024}:
\begin{subequations} \label{eq_cost}
  \begin{equation} \label{eq_cost_1}
    \funcName{cost}(\testc, \vect{p},\lst{W},\lst{V},\lst{Q}) = \sum_{m} \funcName{cost}(\lst{M}_m,\lst{W},\lst{V},\lst{Q}), \mkern22.5mu
  \end{equation}
  \begin{equation} \label{eq_cost_2}
    \begin{split}
    &\funcName{cost}(\lst{M}_m,\lst{W},\lst{V},\lst{Q}) %& \sum_{\testcs_s \in \lst{M}_m} \sum_{j=0}^{j\iEnd\iChildSec{s}} \|\reachY(t\iChildSec{s}[j])\| \,\Delta t\iChildSec{s}[j]. \nonumber \\
    = \mkern-11mu \sum_{\testcs_s \in \lst{M}_m} \mkern-5mu \sum_{j=0}^{j\iEnd\iChildSec{s}-1} \mkern-5mu \Delta t\iChildSec{s}[j] \vect{\beta}^\mathsf{T} \matr{F}[j]
    \begin{bmatrix}
      {\vect{\alpha}_{Q_q}} \\ \vect{\alpha}_W \\ \vect{\alpha}_V 
    \end{bmatrix}, \\
  &\matr{F}[j] = \Big[\lvert \matr{C}\matr{E}_1[j]\matr{G}_{Q_q} \rvert \;\; \lvert \matr{C}\matr{E}_2[j]\matr{G}_W \rvert \;\; \lvert\matr{G}_V\rvert \Big], 
  \end{split}  
\end{equation}
\end{subequations}
where the outer loop cost \eqref{eq_cost_1} is the sum of the inner loop costs \eqref{eq_cost_2} used in linear programming.
The weight vector $\vect{\beta} \in \mathbb{R}^{o}$ determines the importance of each output variable,
which will be shown in Sec. \ref{subsec_exp1}.
Obviously, the computation of $\matr{F}[j]$ can reuse the terms computed for \eqref{eq_linearContain_1}.
Alternatively,
we can choose a small integration step $\Delta t$ (e.g., \SI[]{1}{ms}) for accuracy
when the time step size of the test case is large.

%%%%%%%%%%%%%%%%%%%%%%%%%%%%%%%%%%%%%%%%%%%%%%%%%%%%%%%%%%%%%%%%%%%%%%%%%%%%%%%%
\section{Experimental Results} \label{sec_exp}
\begin{figure}[h]
  \centering
  \includegraphics[width=1\columnwidth]{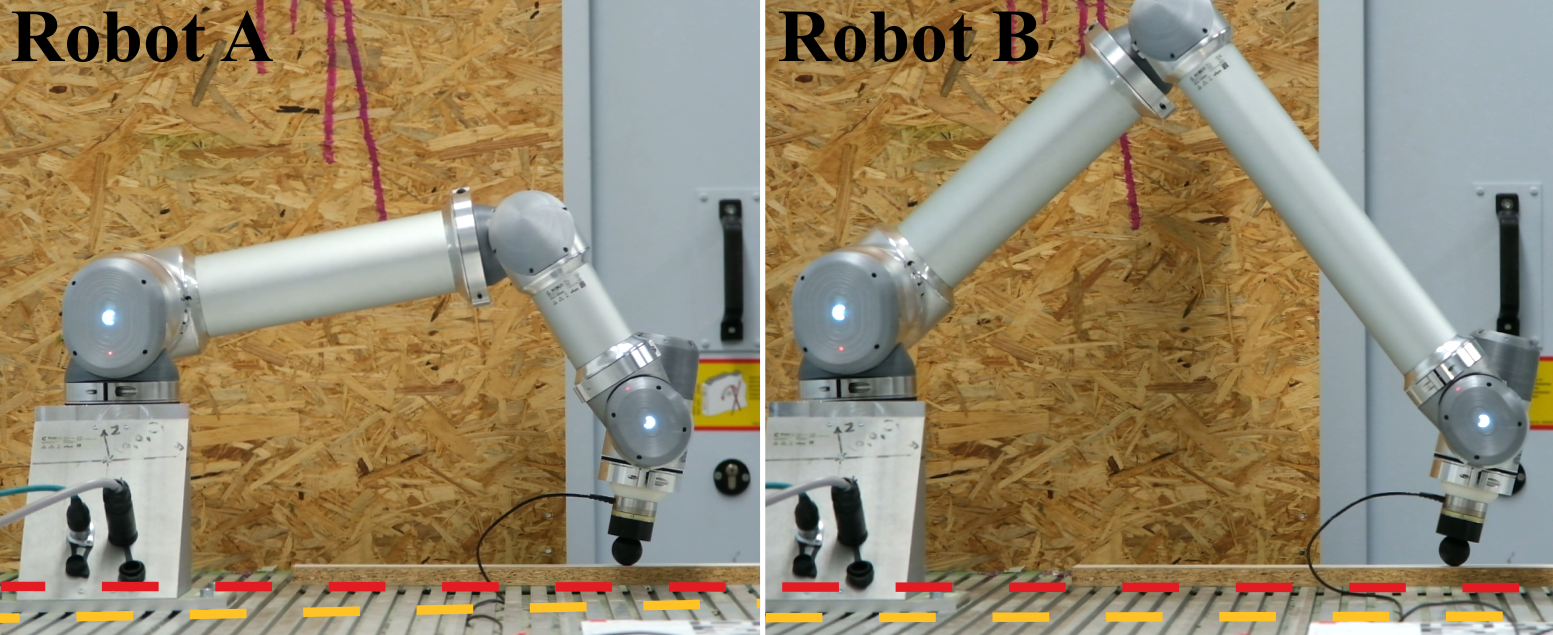}
  \caption{Two 3-DOF robots hitting a wood board.
  The collision force tilts the robot mounting (yellow line) with respect to the board (red line). 
  }
  \label{fig_robot}
  \vspace{0em}
\end{figure}

We test our approach using the scenario in \mbox{Sec. \ref{sec_sys}} with two distinct 3-DOF planar robots:
robot A (maximum reach \SI[]{76}[]{cm}) and robot B (maximum reach \SI[]{104}[]{cm}),
both constructed with RobCo modules as shown in \mbox{Fig. \ref{fig_robot}.}
For the experiments, we specify seven input sequences $\inputU_{v_h}[\cdot]$ of \SI{2}{s} 
to hit a rigid plane respectively with vertical speeds \mbox{$v_h = 0.25, 0.225, 0.2, 0.175, 0.15, 0.125, {0.1}$} and then leave the plane;
the translation and rotation in other dimensions are randomly generated with limits \mbox{$\left|v_x\right| \leq 0.1$}, $\left|\omega_y\right| \leq \pi/7$.
Each robot follows each trajectory $\inputU_{v_h}[\cdot]$ five times, collecting test cases $\testc_{A,v_h}, \testc_{B,v_h}$ respectively for robot A and B;
each test case contains 2000 measurements collected at \SI[]{1}[]{kHz}.

Using the methods in \mbox{Sec. \ref{sec_syn}}, we synthesize several reachset conformant models
and check the measurements against the reachable sets to show the effectiveness of our approach.
Before discussing the results,
we list the synthesized models in \mbox{Tab. \ref{table_params}} and \mbox{Tab. \ref{table_WV}}:
\mbox{Tab. \ref{table_params}} lists the model parameters obtained with nonlinear programming 
while the initial guess is extracted from the controller gains \eqref{eq_fr} or estimated in advance using basic optimization techniques.
\mbox{Tab. \ref{table_WV}} details the sets $\set{W}_2, \set{V}_2, \set{Q}_2$ of $\locat_2$ (\textit{contact}) and $\tuple{Q}_2$ (transition to $\locat_2$).
%Besides the center vector, we list the diagonal elements of the generators while the generator templates are simply identity matrices.
All computations are performed in MATLAB on a \SI[]{2.5}[]{GHz} i9 processor with \SI[]{32}[]{GB} memory within CORA \cite{althoffIntroductionCORA20152015}.

\begin{table}[h]
  \vspace{0em}
  \centering
  \caption{Identified model parameters $\vect{p}$.}
  \label{table_params}
  \setlength{\tabcolsep}{4pt}
  \begin{tabular}{cccccccc}
    \toprule
    model & $h_1$ & $h_2$ & $m_r$ & $k_r$ & $d_r$ & $k_e$ & $d_e$   \\
    \midrule
    $\tuple{H}_{n,A,2}$ & -0.1281 & -0.1288 & 11.0 & 429.7 & 990.4 & 36479.6 & 171.1 \\
    $\tuple{H}_{n,A,1}$ & -0.1284 & -0.1284 & 10.0 & 399.8 & 1000.0 & 36487.4 & 175.4 \\
    $\tuple{H}_{e,A,1}$ & -0.1284 & -0.1283 & 10.0 & 400.0 & 1000.0 & 36487.4 & 175.5 \\
    $\tuple{H}_{n,AB,1}$ & -0.1270 & -0.1269 & 5.4 & 400.7 & 999.8 & 24865.2 & 242.3 \\
    \bottomrule
  \end{tabular}
  \vspace{0em}
\end{table}

\begin{table*}[t]
  \setlength{\tabcolsep}{2pt}
  \renewcommand{\arraystretch}{0.8}
  \centering
  \setlength{\extrarowheight}{0pt}
  \addtolength{\extrarowheight}{\aboverulesep}
  \addtolength{\extrarowheight}{\belowrulesep}
  \setlength{\aboverulesep}{1pt}
  \setlength{\belowrulesep}{0pt}
  \caption{Identified sets $\set{Q}_2, \set{W}_2, \set{V}_2$ (zonotopes) using identity matrices as generator templates.}
  \label{table_WV}
  \begin{tabular}{cccccc@{\hspace{4\tabcolsep}}ccccc@{\hspace{4\tabcolsep}}cccc}
    \toprule
    & \multicolumn{5}{c}{\textbf{elements of $\vect{c}_{Q_2}$}} & \multicolumn{5}{c}{\textbf{elements of $\vect{c}_{W_2}$}} & \multicolumn{4}{c}{\textbf{elements of $\vect{c}_{V_2}$}} \\
    \cmidrule(r){2-6} \cmidrule(lr){7-11} \cmidrule(l){12-15}
    model &  $p_z$ & $v_z$ & $f_e$ & $p_x$ & $\theta_y$ & $p_z$ & $v_z$ & $f_e$ & $p_x$ & $\theta_y$ & $p_z$ & $f_e$& $p_x$ & $\theta_y$ \\ 
    \midrule
    $\tuple{H}_{n,A,2}$ &3.5e-03 & 0.025 & 64.71 & 6.0e-04 & 6.0e-03 & -4.9e-03 & -8.1e-12 & -2.102 & -5.5e-04 & -5.5e-03 & -3.0e-03 & -43.69 & 4.5e-05 & -3.0e-03 \\                                                                                                  
    $\tuple{H}_{n,A,1}$ &7.8e-04 & 0.077 & -8.511 & 6.2e-04 & 6.2e-03 & -8.1e-04 & -2.2e-10 & -0.325 & -5.5e-04 & -5.5e-03 & -9.9e-04 & 32.19 & 2.9e-05 & -3.2e-03 \\
    $\tuple{H}_{e,A,1}$ &1.1e-03 & 0.144 & -8.689 & 1.6e-03 & 6.2e-03 & -1.2e-03 & 1.8e-08 & -0.497 & -1.4e-03 & -5.5e-03 & -1.4e-03 & 53.29 & -9.7e-04 & -3.7e-03 \\
    $\tuple{H}_{n,AB,1}$ &3.3e-03 & -0.088 & 7.036 & 4.8e-03 & 0.011 & -2.3e-03 & -2.4e-09 & -0.936 & -4.3e-03 & -9.7e-03 & -1.7e-03 & 19.46 & -3.8e-03 & -8.4e-03 \\
    \midrule
    & \multicolumn{5}{c}{\textbf{elements of $\vect{\alpha}_{Q_2}$}} & \multicolumn{5}{c}{\textbf{elements of $\vect{\alpha}_{W_2}$}} & \multicolumn{4}{c}{\textbf{elements of $\vect{\alpha}_{V_2}$}} \\                     
    \cmidrule(r){2-6} \cmidrule(lr){7-11} \cmidrule(l){12-15}
    model &  $p_z$ & $v_z$ & $f_e$ & $p_x$ & $\theta_y$ & $p_z$ & $v_z$ & $f_e$ & $p_x$ & $\theta_y$ & $p_z$ & $f_e$& $p_x$ & $\theta_y$ \\ 
    \midrule
    $\tuple{H}_{n,A,2}$ & 4.2e-09 & 0.164 & 72.03 & 5.7e-04 & 1.9e-03 & 8.0e-11 & 1.5e-08 & 3.0e-06 & 9.4e-11 & 1.6e-10 & 1.9e-04 & 88.87 & 5.7e-04 & 1.9e-03 \\                                                                                                     
    $\tuple{H}_{n,A,1}$ &9.2e-08 & 0.109 & 0.167 & 5.7e-04 & 1.9e-03 & 1.7e-09 & 3.3e-07 & 6.2e-05 & 8.0e-08 & 1.2e-07 & 2.7e-04 & 153.5 & 5.7e-04 & 1.9e-03 \\ 
    $\tuple{H}_{e,A,1}$ &1.2e-08 & 0.147 & 0.429 & 6.1e-04 & 2.1e-03 & 2.9e-10 & 5.7e-08 & 1.1e-05 & 1.7e-11 & 1.5e-11 & 2.9e-04 & 239.6 & 6.1e-04 & 2.1e-03 \\  
    $\tuple{H}_{n,AB,1}$ &1.8e-04 & 9.9e-09 & 0.575 & 2.1e-03 & 4.5e-03 & 2.7e-11 & 2.4e-09 & 5.9e-07 & 4.4e-08 & 4.3e-08 & 1.3e-03 & 152.5 & 2.1e-03 & 4.5e-03 \\       
    \bottomrule 
  \end{tabular}
  \vspace{0em}
\end{table*}

\subsection{Modeling the Normal Working Condition of Robot A} \label{subsec_exp1}
\begin{figure}[t]
  \centering
  \includegraphics[width=1\columnwidth]{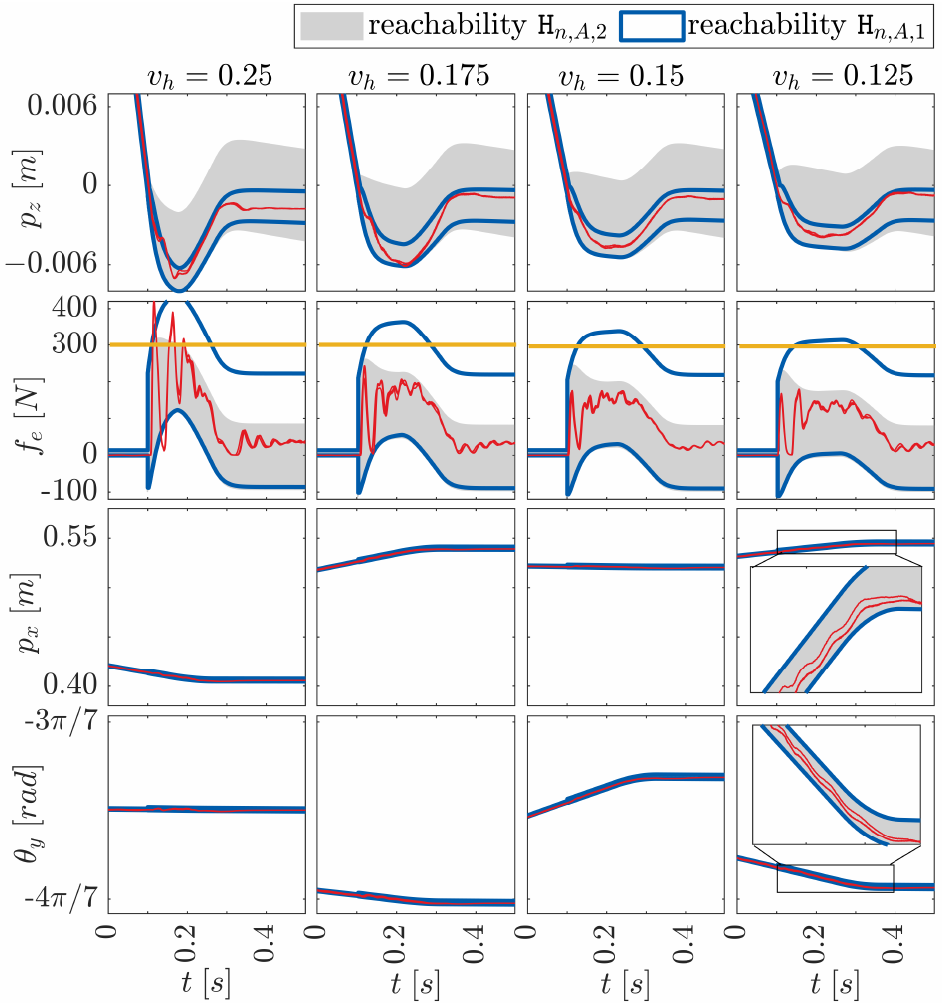}
  \caption{The reachable sets of the identified models $\tuple{H}_{n,A,\beta_f}$ given untested trajectories $\inputU_{0.25}[\cdot], \inputU_{0.175}[\cdot], \inputU_{0.15}[\cdot], \inputU_{0.125}[\cdot]$ (left to right).
  Each trajectory is executed five times on robot A and the measurements are depicted in red lines.
  The yellow lines represent an example safety threshold $f_e = 300$.
  Please note that we offset $\vect{p}_z$ by $-h_1$ of $\tuple{H}_{n,A,1}$ for clearer visualization.
  }
  \label{fig_result}
  \vspace{0em}
\end{figure}

Both robots are built to work with $v_h \leq 0.2$,
while collisions at higher speeds can generate forces over \SI[]{300}[]{N}, potentially damaging mechanical parts.
We first build conformant models for the normal working condition of robot A.
Using recordings \mbox{$\testc_{A,0.1}, \testc_{A,0.2}$} (10 test cases), we synthesize models $\tuple{H}_{n,A,\beta_f}$, 
where \mbox{$\beta_f=1,2$} is the weight of contact force in the defined weight vector \mbox{$\beta=[k_e, \beta_f, k_e, k_e]^\mathsf{T}$},
and $k_e$ is the initial guess of the contact stiffness.
We compute the reachability of the models using CORA given four untested trajectories
and check all measurements against the reachable sets, as shown in \mbox{Fig. \ref{fig_result}}.

It can be seen that both models capture all measurements for $v_h = 0.175, 0.15, 0.125$, although they are unseen to the synthesis procedure.
Accordingly, with minimal testing, we can create conformant models for a range of working conditions and efficiently verify safety properties.
For example, we can verify if the contact force might exceed the specified safety threshold $f_e = 300$ (yellow lines in \mbox{Fig. \ref{fig_result}}) by checking its intersection with the reachable sets.
The weight vector allows adjusting the model for verifying specific outputs.
In our case, a larger $\beta_f$ tightens the force dimension of the reachable sets while widening others.
Accordingly, $\tuple{H}_{n,A,1}$ helps verify the position, while $\tuple{H}_{n,A,2}$ helps verify the force;
using both leads to overall more precise verification results.
Also, as expected, the exception case $v_h = 0.25$ is not well captured,
as the robot mounting tilts much (see Fig. \ref{fig_robot}) due to the large contact force,
and the caused errors cannot be enclosed when they are unseen to the synthesis procedure.

To synthesize the system described, given 10 test cases with $k\iEnd = 2000$,
the data preprocessing (Sec. \ref{subsec_testCases}) takes \SI[]{1.6}[]{s},
and the identification with linear programming takes \SI[]{806}[]{s} using the generator enclosure method \mbox{\cite[Thm. 3]{lutzowReachsetConformantIdentificationNonlinear2024}}
and \SI[]{229}[]{s} using the halfspace method \mbox{\cite[Thm. 1]{liuGuaranteesRealRobotic2023}}, with similar identification results.
The halfspace method is faster with fewer linear programming variables, as the often-used interior point algorithms \cite{vandenbrandDeterministicLinearProgram2020} scale polynomially with the number of variables.
However, without the under-approximation \eqref{eq_underapproxW} and for higher-dimensional systems,
the generator method is preferable as it scales better with time horizon and system dimension \cite[Sec.~\ref{sec_exp}]{lutzowReachsetConformantIdentificationNonlinear2024} due to the fewer linear constraints.
In practice, solving the outer loop using nonlinear programming often requires hundreds of evaluations of the inner loop;
obtaining a good initial guess for $\vect{p}$ based on solely the outer loop can significantly reduce iterations.
Alternatively, we downsample the test cases when synthesizing model parameters; the inner loop only consumes seconds given a test case with $k\iEnd = 1000$.
After obtaining $\vect{p}$, we run the inner loop again with the original test cases to formally enclose all measurements.

\begin{figure}[t]
  \centering
  \includegraphics[width=1\columnwidth]{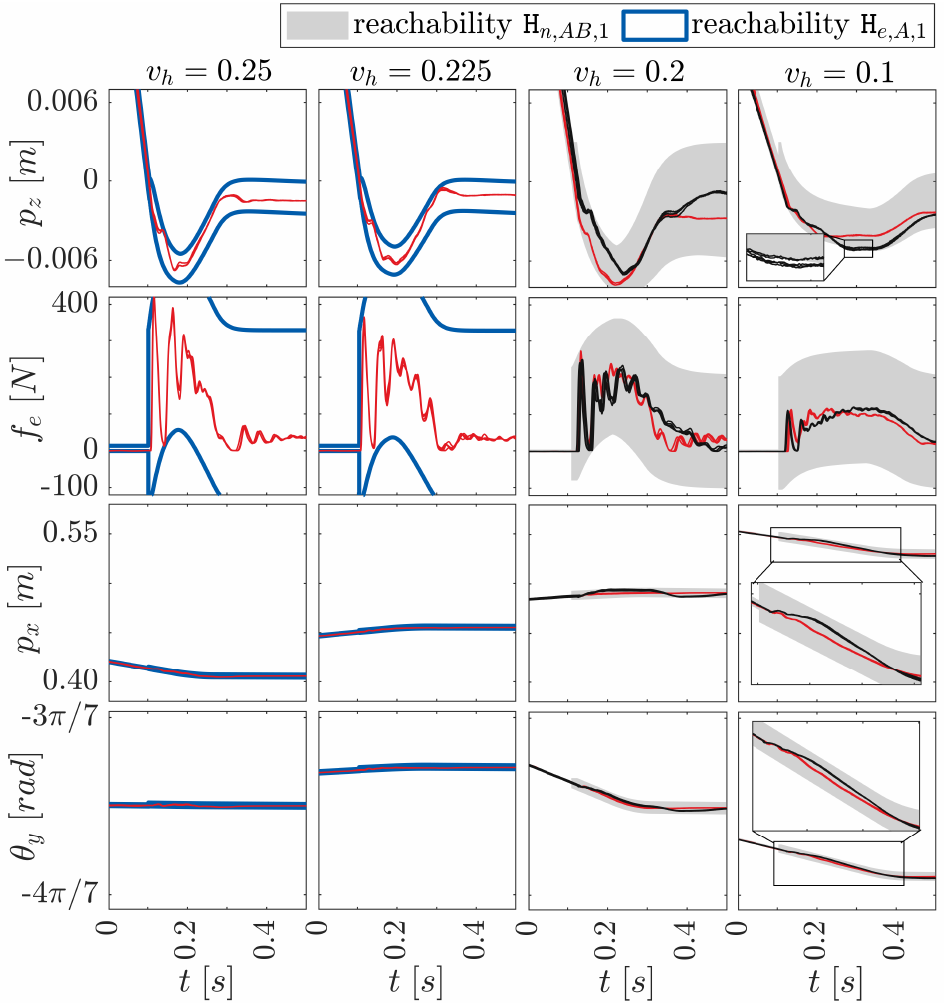}
  \caption{The reachable sets of the identified model $\tuple{H}_{e,A,1}$ given trajectories \mbox{$\inputU_{0.25}[\cdot], \inputU_{0.225}[\cdot]$} and the reachable sets of $\tuple{H}_{n,AB,1}$ given trajectories \mbox{$\inputU_{0.2}[\cdot], \inputU_{0.1}[\cdot]$} (left to right).
  Each trajectory is executed five times on both robots. The measurements of robot A are depicted in red lines, and the measurements of robot B are depicted in black.
  Please note that we offset $\vect{p}_z$ by $-h_1$ of the used model for clearer visualization.
  }
  \label{fig_result_2}
  \vspace{0em}
\end{figure}

\subsection{Enclosing Exception Cases and Other Robots with Little Testing Effort} \label{subsec_exp2}
While testing exception cases with $v_h > 0.2$ is risky,
one option is to test an edge case a few times and feed the data to conformance synthesis
along with the test cases of normal working conditions,
as the latter should already capture most system uncertainties.
As an example, we add one test case $\testc_{A,0.25}$ to the 10 test cases $\testc_{A,0.1},\testc_{A,0.2}$ used in \mbox{Sec. \ref{subsec_exp1}}
and obtain model $\tuple{H}_{e,A,1}$.
The reachable sets of $\tuple{H}_{e,A,1}$ are plotted together with the measurements in Fig. \ref{fig_result_2} (the left two columns).
It can be seen that the measurements are indeed captured by the synthesized model.

Similarly, we can extend the model for other robots with little additional testing, which is useful in flexible manufacturing environments.
By adding only one test case $\testc_{B,0.2}$ collected from robot B to the test cases $\testc_{A,0.1},\testc_{A,0.2}$ of \mbox{robot A},
we obtain model $\tuple{H}_{n,AB,1}$ for the normal working condition of both robots.
We feed the trajectories $\inputU_{0.1}[\cdot], \inputU_{0.2}[\cdot]$ to the model, and the measurements of both robots are checked against the resulting reachable sets in Fig. \ref{fig_result_2} (the right two columns).
It can be seen that
the model captures the differences and encloses most of the measurements with the reachable sets,
although the behavior of robot B is very different from robot A.
The only exception is within the position $p_z$ for \mbox{$v_h = 0.1$}, which is enlarged in Fig. \ref{fig_result_2}:
Some measurements of robot B are slightly outside (distance $<{0.0002}$) of the reachable set for a short duration (\SI{0.05}{s}),
which can be easily captured by feeding more test cases or manually bloating the reachable set for such an extreme case with a different system and very little testing.

%%%%%%%%%%%%%%%%%%%%%%%%%%%%%%%%%%%%%%%%%%%%%%%%%%%%%%%%%%%%%%%%%%%%%%%%%%%%%%%%
\section{Conclusions}
We synthesize reachset-conformant hybrid models for delivering safety properties to real robotic contact tasks.
For the first time,
reachset conformance is established for robotic systems with hybrid dynamics considered,
and the conformance synthesis is done optimally for hybrid systems with each location separately identified using linear programming. 
With a typical contact task scenario,
experiments show the effectiveness and usefulness of our approach:
The synthesized conformant model can well capture untested conditions when the model is given a short recording,
and with a small amount of additional testing, the model can further enclose large uncertainties that were unseen,
such as an unusual behavior or a different robot.
Accordingly, our approach can largely reduce the required testing effort,
which is particularly valuable for conditions that are difficult to test or for flexible manufacturing environments that demand frequent modifications.

%%%%%%%%%%%%%%%%%%%%%%%%%%%%%%%%%%%%%%%%%%%%%%%%%%%%%%%%%%%%%%%%%%%%%%%%%%%%%%%%
\section*{Acknowledgment}
This work was funded in part by the Siemens AG and the Deutsche Forschungsgemeinschaft (DFG, German Research Foundation) - SFB 1608 - 501798263.
The authors thank Bastian Sch{\"u}rmann and Florian Wirnshofer for the valuable discussions regarding the project.

%%%%%%%%%%%%%%%%%%%%%%%%%%%%%%%%%%%%%%%%%%%%%%%%%%%%%%%%%%%%%%%%%%%%%%%%%%%%%%%%
\bibliography{IROS2024_ReachsetConformanceContactTasks}

\end{document}